\documentclass[runningheads]{llncs}

\usepackage[]{accv}

\usepackage{accvabbrv}

\usepackage{graphicx}
\usepackage{array}
\usepackage{booktabs}
 \usepackage{multirow}
\usepackage[accsupp]{axessibility}  

\usepackage[pagebackref,breaklinks,colorlinks,citecolor=accvblue]{hyperref}

\usepackage{orcidlink}

\begin{document}

\title{Region Prompt Tuning: Fine-grained Scene Text Detection Utilizing Region Text Prompt} 

\titlerunning{Abbreviated paper title}

\author{Xingtao Lin\and
Heqian Qiu\and
Lanxiao Wang\and
Ruihang Wang\and Linfeng Xu \and Hongliang Li}

\authorrunning{F.~Author et al.}

\institute{University of Electronic Science and Technology of China, Chengdu, China
}

\maketitle

\begin{abstract}
Recent advancements in prompt tuning have successfully adapted large-scale models like Contrastive Language-Image Pre-trained (CLIP) for downstream tasks such as scene text detection. Typically, text prompt complements the text encoder's input, focusing on global features while neglecting fine-grained details, leading to fine-grained text being ignored in task of scene text detection. In this paper, we propose the region prompt tuning (RPT) method for fine-grained scene text detection, where region text prompt proposed would help focus on fine-grained features. Region prompt tuning method decomposes region text prompt into individual characters and splits visual feature map into region visual tokens, creating a one-to-one correspondence between characters and tokens. This allows a character matches the local features of a token, thereby avoiding the omission of detailed features and fine-grained text. To achieve this, we introduce a sharing position embedding to link each character with its corresponding token and employ a bidirectional distance loss to align each region text prompt character with the target ``text''. To refine the information at fine-grained level, we implement character-token level interactions before and after encoding. Our proposed method combines a general score map from the image-text process with a region score map derived from character-token matching, producing a final score map that could balance the global and local features and be fed into DBNet to detect the text. Experiments on benchmarks like ICDAR2015, TotalText, and CTW1500 demonstrate RPT impressive performance, underscoring its effectiveness for scene text detection.

  \keywords{Region prompt tuning \and CLIP \and Character-token level }
\end{abstract}
\begin{figure}[t]
\centering
\subfloat[Current prompt tuning methods focus on the general feature of the whole image.]{\includegraphics[width=0.46\textwidth]{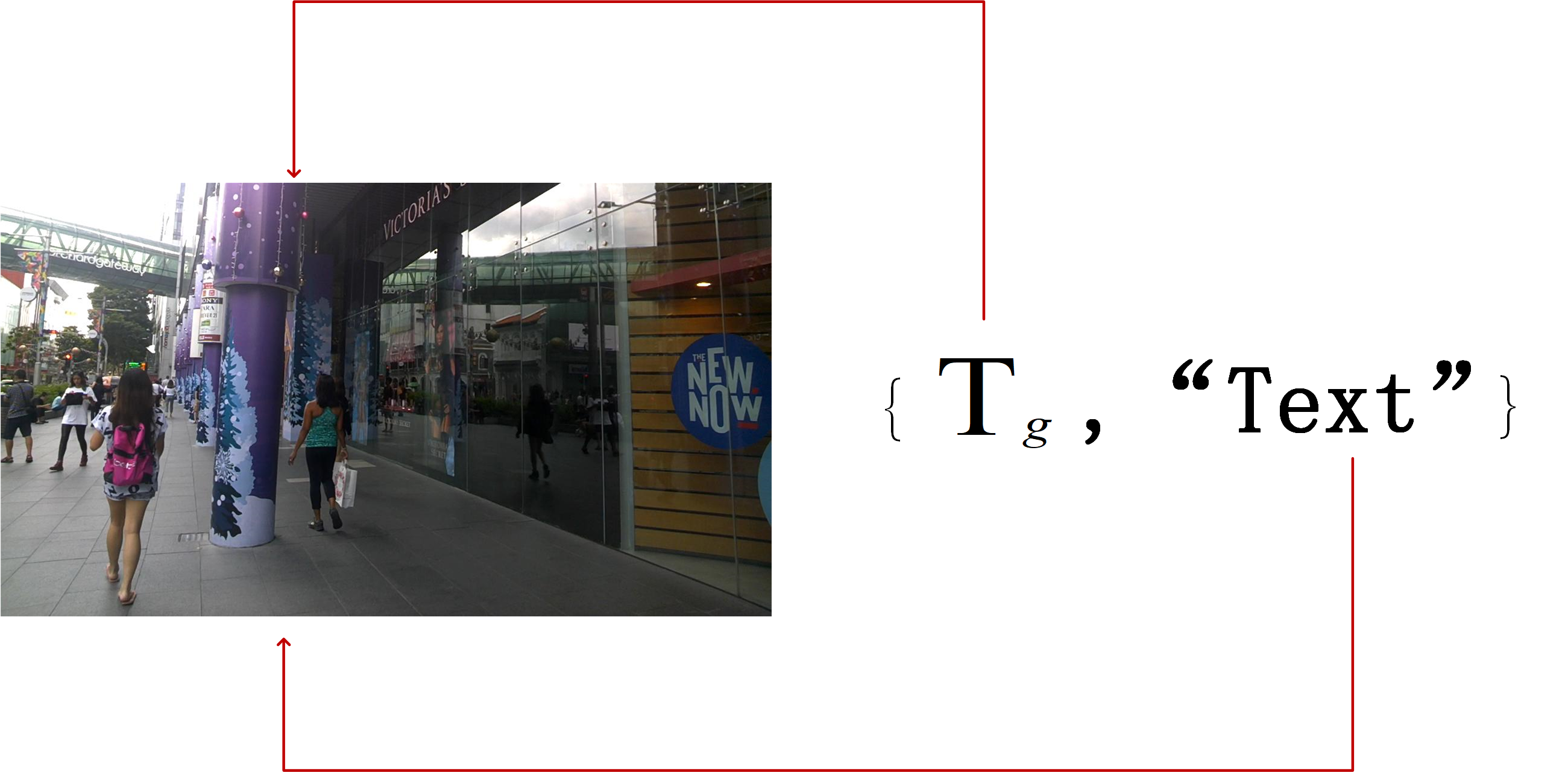}
\label{base}}
\hfill
\subfloat[Our proposed method region prompt tuning, where a region text prompt character corresponds one-to-one with a region visual token. ]{\includegraphics[width=0.50\textwidth]{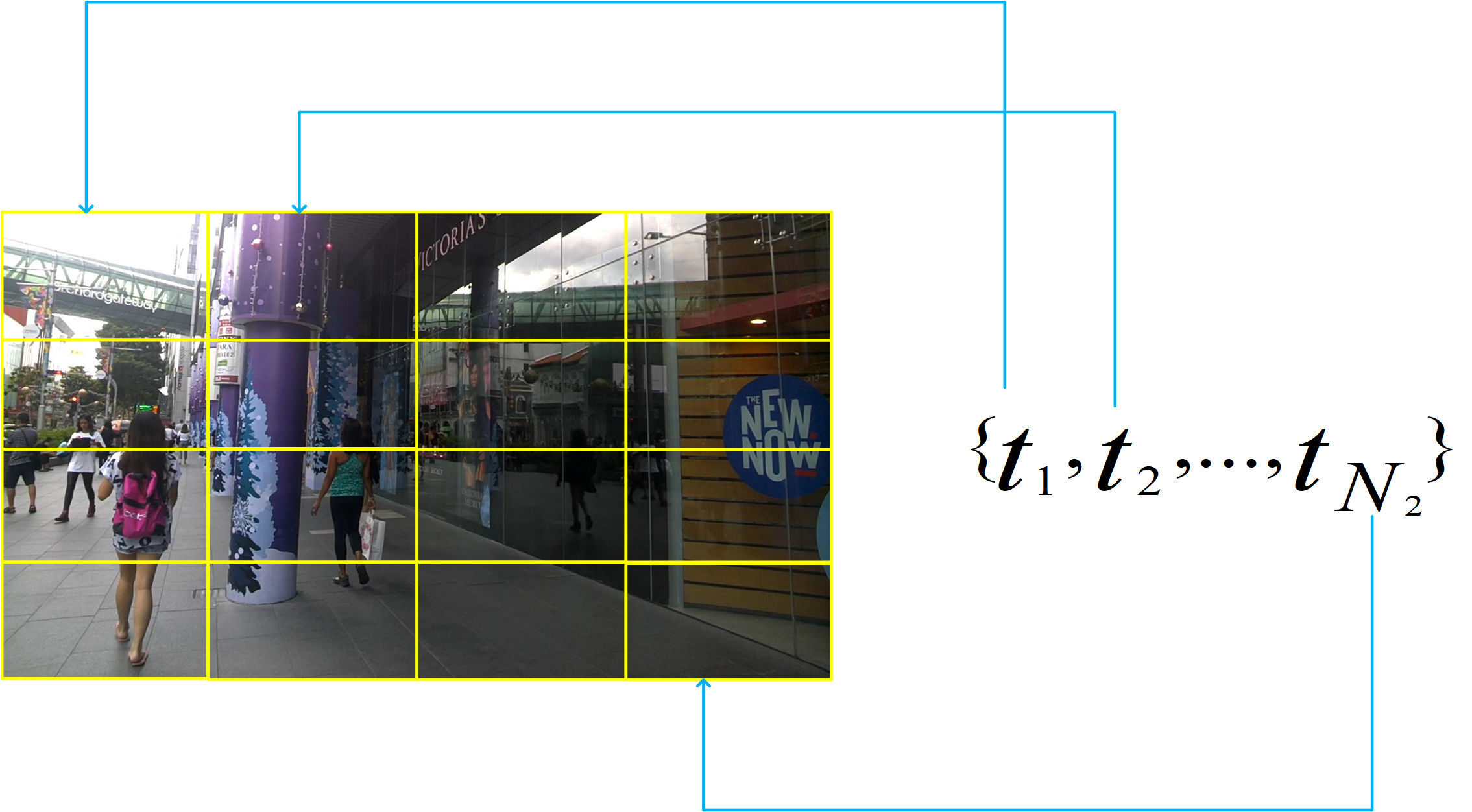}
\label{ours}}
\caption{Comparison between our proposed method and the current paradigms of the prompt learning. ``$T_g$'' stands for the text prompt of the current paradigms of the prompt learning, while ``$t$'' represents for our proposed region text prompt character.}
\label{detail}
\end{figure} 

\section{Introduction}
\label{sec:intro}
Scene text detection is a crucial task in image processing, focusing on locating characters, words, and sentences within random scenes in raw images. Its applications span various industries, including automated document processing, scene understanding, autonomous driving, and virtual reality. Therefore, by ensuring the accuracy of the text detection, the further task such as text recognition would be expanded and promoted further.
\par
Since the Contrastive Language–Image Pre-training (CLIP) \cite{CLIP} is a robust large-scale model for bridging the gap between the natural languages and the images, the CLIP model is widely used for the downstream visual task, also could be introduced into our core task, scene text detection. Pre-trained on millions of image-text pairs, the merits of the CLIP model lies in its generalization ability, leading it to provide a good performance on wide range of the tasks without too many re-training on the task-specific data.
\par
Based on the CLIP model, the techniques of prompt tuning has widely involved in transferring the original CLIP model to the specific tasks better. With the specially designed prompt, the alignment between the textual descriptions and the images tends to be maximized to enhance the performance of the original CLIP model. The prompt related to the specific tasks could be capable of making the original CLIP model to the downstream task more quickly.  Additionally, prompt learning is computationally efficient, as it fine-tunes large-scale models without the need to retrain them entirely, making it an effective approach for optimizing performance.
\par
However, there still exists several shortcomings in the current paradigms of the prompt tuning, preventing it from fully leveraging the powerful capacities of the CLIP model. Since the CLIP model always allows itself to capture broad semantic relationships between the images and the texts pre-trained on millions of annotations, the powerful generalization ability, which could make the CLIP model as a double-edged sword, leads the CLIP model to focus on more prominent and global features, while ignore some detailed information, to be specific, some fine-grained text features in the task of text detection. However, the current prompt tuning paradigms, for example, like CoOp \cite{coop} and CoCoOp \cite{cocoop}, always treat the text prompt as an general implementation of the text input, with no specific and unclear meanings, as illustrated in Fig. 1 (a), and put this text prompt for the further matching with the whole image input. In fact, the current prompt-tuning methods are helpless against the overlook of the fine-grained features in the original CLIP model. 
\par
To address the aforementioned disadvantages leading overlook of the fine-grained text, we introduce a region text prompt, which consists of region text prompt characters. The visual feature map generated by the ResNet-50 \cite{Res} backbone could also be decomposed into region visual tokens, each corresponding to a region text prompt character. This establishes a one-to-one character-to-token correspondence, where each character matches the fine-grained features of its corresponding region visual token, as shown in Fig. 1 (b), rather than supplementing the original text input, as shown in Fig. 1 (a). In this paper, we propose a region prompt tuning for fine-grained scene text detection, called RPT. 
\par
In our proposed RPT framework, two categories of text prompts coexist in parallel. One is the general text prompt, working as a implementation of the fixed word embedding ``text''. The other is the region text prompt, composed of individual characters, designed to capture fine-grained local features at the region visual token level. To establish a character-token correspondence between the region text prompt and region visual token, a mechanism of sharing position embedding has developed. Meanwhile, character-token level interactions are introduced to refine the characters and their corresponding tokens through Transformer decoder both before and after encoding. Additionally, a bidirectional distance loss is applied to align the region text prompt with the detection target ``text'' and to bring the general text prompt closer to the region text prompt. Parallel to the traditional image-text matching, which generates a general score map, we adopt a character-token matching method to match individual characters with their corresponding tokens to obtain a region score map. We then employ feature enhancement and fusion methods to combine the general and region score maps, deriving the final score map to balance the global and region features. This final score map is fed into the detection head of DBNet \cite{dbn} to produce the detection result. Our contributions are summarized as follows:
\begin{itemize}
\item[$\bullet$]
Different from current prompt tuning paradigms, our RPT framework introduces a region text prompt to capture fine-grained features at the character-token level. By sharing position embeddings and facilitating interactions between characters and corresponding tokens, we generate a region score map using a character-token matching method to reinforce the local features, in order to avoid neglecting the fine-grained text in the token level.        
\item[$\bullet$]
A bidirectional distance is developed to focus the region text prompt on the detection target ``text'', and make the general text prompt to pay attention to some fine-grained features bidirectionally.

\item[$\bullet$]
A general score map and a region score map is utilized parallel, serving for the feature enhancement and feature fusion to balance the global features and local features.
\end{itemize}

\section{Related Work}
Recent advancements in scene text detection have seen a shift from traditional deep learning-based methods like SPCNet \cite{spc}, Mask TextSpotter \cite{mask}, TextBoxes \cite{box}, and TextSnake \cite{snake} to more sophisticated approaches. The CLIP model by OpenAI revolutionized multimodal learning by aligning images and text in a shared embedding space through contrastive learning, excelling in tasks like text detection and zero-shot learning, with Yu et al. \cite{TCM} introducing it into scene text detection. Prompt learning has further optimized pre-trained models; Zhou et al. \cite{coop} introduced manually designed and fully learnable prompts for classification, Rao et al. \cite{denseclip} explored text prompts, and Jia et al. \cite{vpt} applied visual prompts within CLIP’s ViT \cite{vit} structure, while Yu et al. \cite{TCM} integrated these prompts into text detection tasks. Finally, DBNet (Differentiable Binarization Network) \cite{dbn} has become a state-of-the-art method, utilizing a differentiable binarization technique to enhance text detection in complex environments.

\section{Methodology}

In this paper, we present a framework of fine-grained text detection called RPT based on proposed region text prompt. We introduce general and region text prompt, establish character-token correspondence by sharing position embedding, import bidirectional distance loss and character-token interaction, and generate the final detection result from the score map derived from feature enhancement and fusion. The detailed framework is shown in Fig. 2.  
\begin{figure*}
\centering
\includegraphics[width=\textwidth]{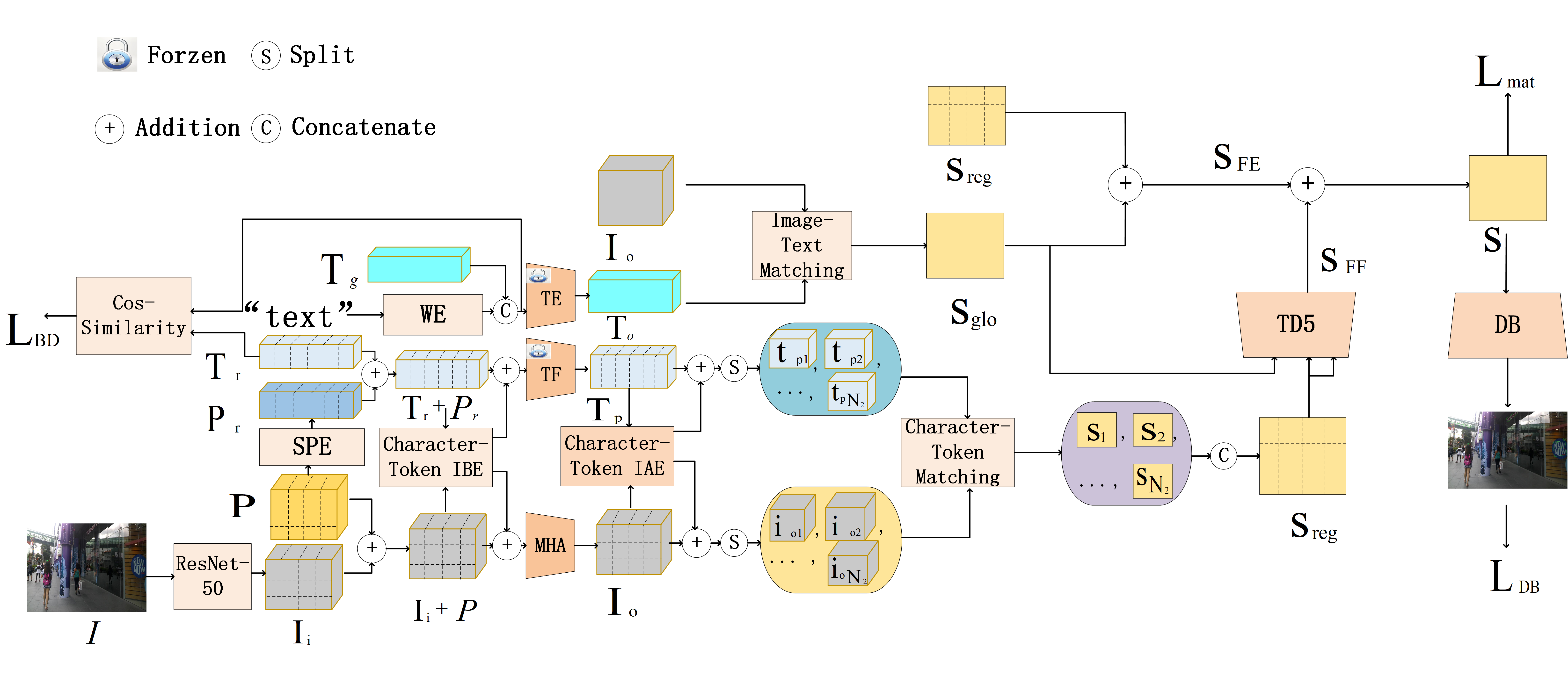}
\caption{The detailed framework of RPT, where ``TD'' is short for Transformer Decoder, ``SPE'' is short for the mechanism of Sharing Position Embedding, ``TF'' is short for the mechanism of Transformer, ``MHA'' is short for the mechanism of Multi-Head Attention, and ``TE'' is short for Text Encoder. Meanwhile, ``IAE'' and ``IBE'' represent for Interaction Before and After Encoding, ``DB'' is short for DBNet.}
\label{figure}
\end{figure*}
\subsection{Definition of Two Types Learnable Text Prompt}
   Unlike the traditional prompt-tuning paradigm like CoOp \cite{coop} within the CLIP framework, the learnable text prompt we designed can be split into two distinct parts: the general text prompt $T_g \in \mathbb{R}^{{1}\times{N_1}\times{C^{\prime}}} $ and the region text prompt $T_r\in \mathbb{R}^{{1}\times{N_2}\times{C^{\prime}}}$, where $N_1$ stands for the length of $T_g$, and $N_2$ stands for the length of $T_r$, while the dimension of text prompt and word embedding $C^{\prime}$ is set to 512. Since the class to be detected in the task of text detection is only 1, the fixed word embedding $T_f \in \mathbb{R}^{{1}\times{N_f}\times{C^{\prime}}}$ is generated from the single word ``text'' through the word embedding module, calculated as follows: \begin{equation} T_f=WordEmbedding(``text"), \end{equation}
    where $N_f$ is the length of the fixed word embedding. Then, the general prompt $T_g$, which is designed as an implementation of $T_f$ for capturing the global features that fixed word embedding $T_f$, is placed after $T_f$ to form the input of text encoder. The input is called $T_i\in \mathbb{R}^{1 \times{(N_f+N_1)}\times{C^{\prime}}}$, expressed as below: \begin{equation} T_i=[T_f,T_g].\end{equation} Text encoder, from the CLIP framework with all its parameters frozen, encodes the text input  $T_i$ into the final text embedding $T_o\in \mathbb{R}^{1\times{(N_f+N_1)}\times{C}}$, expressed as below: \begin{equation} T_o=TextEncoder(T_i), \end{equation} where C is the dimension of text embedding set to 1024.
\par

Parallel to the text encoding process,  the visual branch uses the ResNet-50 \cite{Res} image encoder from the original CLIP architecture to extract the visual feature map $I_i\in \mathbb{R}^{\frac{H}{d}\times{\frac{W}{d}}\times{C^{\prime\prime}}}$ from the original image input $I\in \mathbb{R}^{H \times {W} \times{3}}$.
This is achieved through the ResNet-50 backbone network, followed by an AttentionPooling2d layer to obtain the final visual embedding $I_o\in \mathbb{R}^{\frac{H}{d}\times{\frac{W}{d}}\times{C}}$.  The process is expressed in two steps as below:

\begin{equation} I_i=ResNet-50(I),  \end{equation}\begin{equation} I_o=AttentionPooling2d(I_i), \end{equation}
where $H$ is the height and $W$ is the width of the input image $I$, $d$ is the downsampling rate set to 32, and $C^{\prime\prime}$ is the dimension of  $I_i$ set to 2048.

In conventional methods like CoOp \cite{coop} and CoCoOp \cite{cocoop}, the general prompt $T_g$
tends to match the visual features $I_i$ of the whole image during the image-text matching process. In contrast, our approach leverages the region prompt $T_r$ to capture fine-grained details within separate regions of the visual feature map $I_i$. In other words, the region prompt $T_r$ is designed to match local features within these fine-grained regions.
To achieve this, we first establish a corresponding relationship between the region text prompt $T_r$ and the visual feature map $I_i$ in character-token level. 
The region text prompt $T_r$ could be split into $N_2$ region text prompt characters, defined as follow:
\begin{equation}
T_r=\{t_1,t_2,...,t_a,...,t_{N_2}  \},    
\end{equation}
where $t_a \in \mathbb{R}^{{1}\times{1}\times{C^{\prime}}}$ means the $a_{th}$ region text prompt character. We select  $N_2$  s a perfect square, which can be expressed as the square of a factor k. To align $N_2$ region text prompt characters with $N_2$ region visual tokens, the visual feature map $I_i$ is divided into k parts along both the height and width dimensions, forming $k^2$ region visual tokens as below:
\begin{equation}
I_i=\{i_1,i_2,...,i_a,...,i_{N_2}\},
\end{equation}
where $i_a \in \mathbb{R}^{\frac{H}{d\cdot k}\times{\frac{W}{d\cdot k}}\times{C^{\prime \prime}}}$  denotes the the $a_{th}$ region visual token, corresponding to the $a_{th}$ region text prompt character $t_a$. This design ensures that each region text prompt character $t_a$ to a specific region visual token $i_a$.
\subsection{Sharing Position Embedding}
\begin{figure}[t]
	\centering
	{\includegraphics[width=0.95\textwidth]{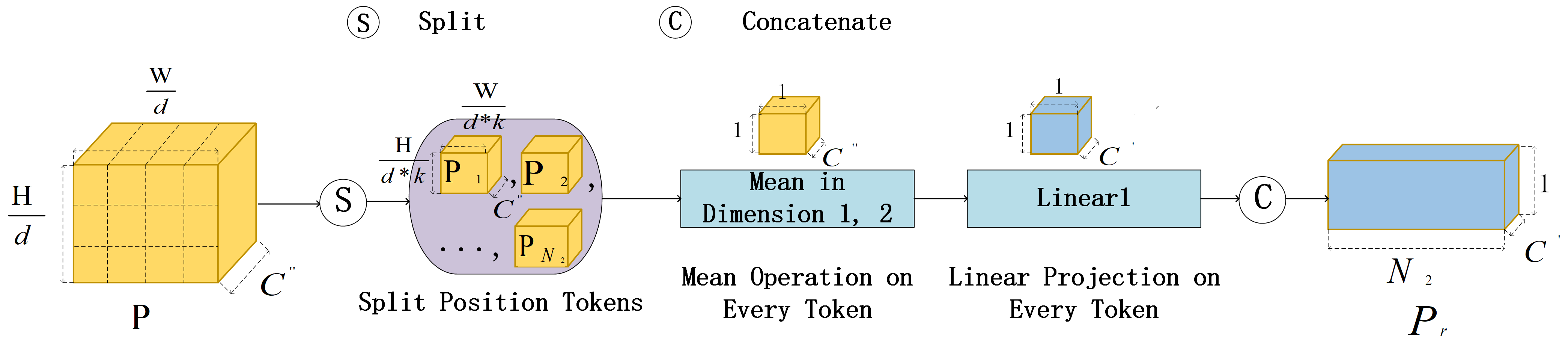}
		\label{base}}
	\caption{The process of sharing position embedding, where every single position character could be obtained from the correspondingly position token one by one.}
\end{figure} 
In the AttentionPooling2d layer, the final visual embedding $I_o$ could be generated from the feature map $I_i$ in two steps. First, the learnable position embedding $P\in \mathbb{R}^{{\frac{H}{d}}\times{\frac{W}{d}}\times{C^{\prime\prime}}}$, is introduced to provide positional information for each pixel in $I_i$. Next, the final visual embedding $I_o$ is obtained through the multi-head attention \cite{att} mechanism:
\begin{equation}I_o=MultiHeadAttention(q,k,v=(I_i+P)). \end{equation}
\par
To keep the output features of $T_f$ and $T_i$ separate, a distinct text encoder, called the prompt encoder, independently encodes the region text prompt $T_t$. Therefore, the text prompt embedding $T_p\in\mathbb{R}^{{1}\times{N_2}\times{C}}$ is the output of encoding $T_r$ through the Prompt Encoder, which inherits almost all the pre-trained parameters of the CLIP text encoder, with the parameters being frozen, except the parameters of the learnable position embedding $P_r \in \mathbb{R}^{{1}\times{N_2}\times{C^{\prime}}}$, as shown below:
\begin{equation} T_p=PromptEncoder(T_r).\end{equation}
\par
In the prompt encoder, the learnable position embedding $P_r$ captures the positional information of each region text character. By adding these parameters to the original region text prompt $T_r$ and feeding them into the Transformer \cite{att} mechanism, the model captures the positional relationships between each region text prompt character as below:
\begin{equation} T_p=Transformer(T_r+P_r).  \end{equation}
\par
The position embedding $P_r$ is designed to capture both the relative positional relationships within each text character and the correspondence between the region text prompt characters and their associated region visual tokens. Therefore, $P_r$ is is generated from the learnable position embedding $P$ in the visual branch instead of inheriting it from the pre-trained CLIP text encoder. This process involves three steps shown in Fig. 3:
\begin{itemize}
\item[$\bullet$] \textbf{Identify Position Tokens}: Identify the position tokens of corresponding region visual tokens in $P$, which shares the same shape as $I_i$. Split $P$ into a set of position tokens as below:
\begin{equation}
P=\{p_1,p_2,...,p_a,...,p_{N_2}\},
\end{equation}
where $p_a \in \mathbb{R}^{\frac{H}{d\cdot k}\times{\frac{W}{d\cdot k}}\times{C^{\prime \prime}}}$  represents the position token of the $a_{th}$ region visual token $i_a$.
\item[$\bullet$]\textbf{Convert Position Tokens to Position Characters}: Convert each position token $P_a$ 
into a corresponding position character $P_{ra}$ as follow:
\begin{equation}
P_{ra}= LN1(Mean(p_a,dim=1,dim=2)),
\end{equation}
where $p_ra \in \mathbb{R}^{{1}\times{1}\times{C^{\prime}}}$, Mean($\cdot$) calculates the average over height and width dimensions on every token, and LN1 is a linear layer projecting $C^{\prime \prime}$ to $C^{\prime}$ in third dimension, transferring a token to a character.  
\item[$\bullet$]
\textbf{Form Final Position Embedding}: Concatenate all position characters to form the final position embedding $P_r$ as below:
\begin{equation}
P_r=\{p_r1,p_r2,...,p_{r{N_2}}\}.
\end{equation}
\end{itemize}
Through these steps, the mechanism of sharing position embedding is introduced.

\subsection{Bidirectional Distance Loss}
There are some defects in the design of region text prompt. Compared to the general prompt $T_g$, which can be seen as an implementation of the fixed text embedding $T_f$, the region text prompt characters might match features from their corresponding region visual tokens that are not closely related to the target ``text.'' Additionally, the general prompt $T_g$ may overlook fine-grained text features in favor of matching the global feature of the visual embedding. To address these issues, a bidirectional distance loss $L_{BD}$ is introduced. This loss helps each region text prompt character focus on the detection target ``text'' while encouraging the general prompt to consider fine-grained text features.  First, a cosine similarity $Sim$ could be calculated between $T_i$ and $T_r$ as below:
\begin{equation} 
Sim=Cos\_Similarity(\mathbf{T_i},\mathbf{T_r})=\frac{ \mathbf{T_i} \cdot \mathbf{T_r}  }{\| \mathbf{T_i} \|_{2} \| \mathbf{T_r} \|_{2}},
\end{equation}
Second, The bidirectional distance loss $L_{BD}$ could be obtain from Sim as follow:
\begin{equation}
L_{BD}=1-sim.    
\end{equation}
By incorporating this bidirectional distance loss, the model balances the focus between global and fine-grained features, improving the overall accuracy and robustness of text detection.
\subsection{Character-token Level Interaction}
\begin{figure}[t]
	\centering
	{\includegraphics[width=0.95\textwidth]{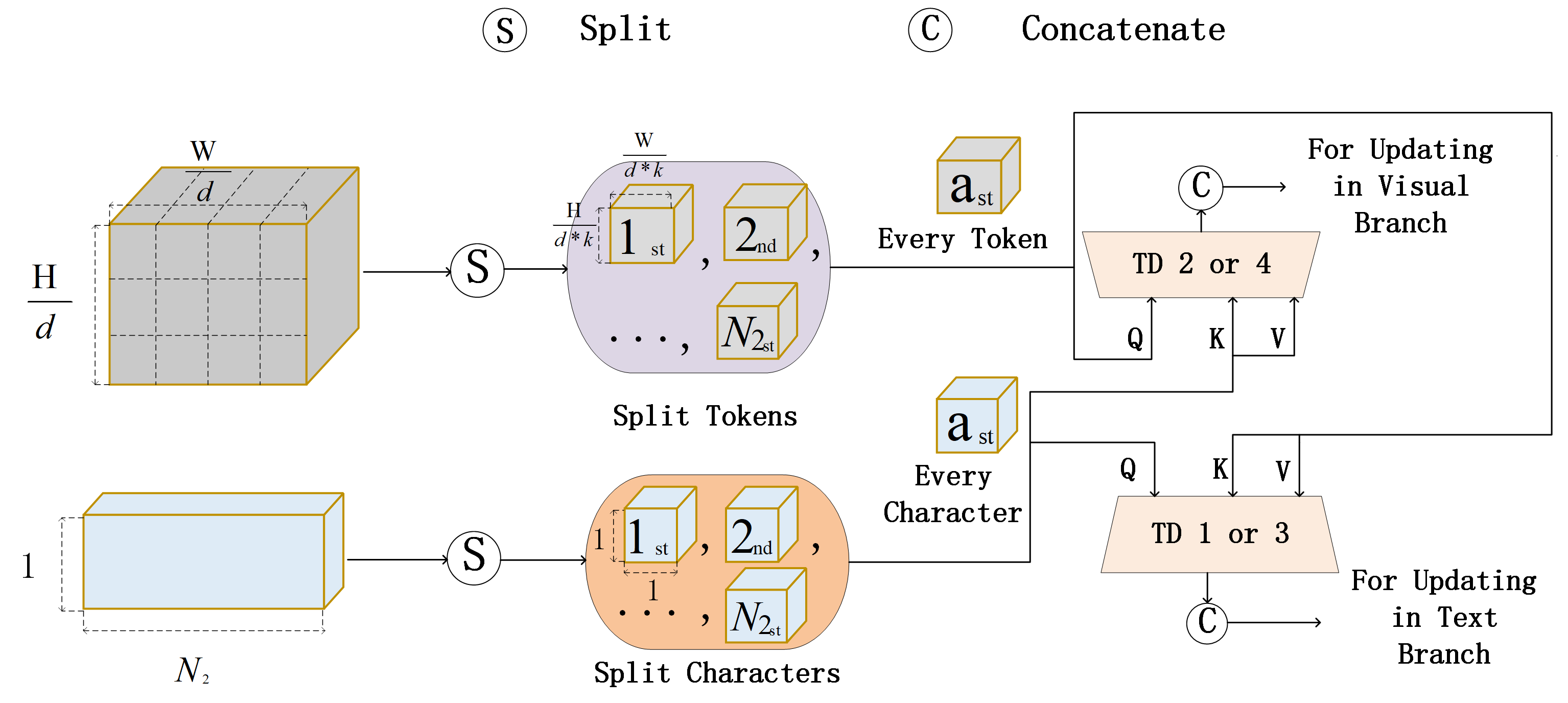}
		\label{base}}
	\caption{Character-token interaction before and after the encoding process, where ``TD'' is short for the Transformer Decoder.}
\end{figure} 
Inspired by DenseClip \cite{denseclip} using Transformer \cite{att} decoder to fuse both text and visual information, we introduce a character-token level interaction both before and after encoding illustrated in Fig. 4, refining text and visual information via 4 Transformer decoders at fine-grained level.
\subsubsection{Pre-Encoding Interaction}
Before encoding the region text prompt $T_r$ and visual feature map $I_i$  with their position embeddings $P_r$ and $P$ respectively, each region text prompt character $t_a$ with position character $p_{ra}$ updated through interaction with the corresponding region visual token $i_a$ and position token $p_a$:
\begin{equation}
t_{a}+p_{ra}=t_{a}+p_{ra}+l_{1}\cdot TransformerDecoder1(q=(t_{a}+p_{ra}),(k,v)=(i_{a}+p_{a})),
\end{equation}
where $l_1$ controls the impact of this interaction. This process updates $T_r+P_r$ at the character level before encoding $T_r+P_r$ through Transformer.
\par
Similarly, in the visual branch, each region visual token $i_a$ with position token $p_a$ is updated through interaction with the corresponding text prompt character $t_a$ and position character $p_ra$: 
\begin{equation}
i_{a}+p_{a}=i_{a}+p_{a}+l_{2}\cdot TransformerDecoder2(q=(i_{a}+p_{a}),(k,v)=(t_{a}+p_{ra})),
\end{equation}
where $l_{2}$ controls the impact of this interaction. This process updates $I_i+P$ at the token level before encoding $I_i+P$ through multi-head attention mechanism.
\subsubsection{Post-Encoding Interaction}
Since the outputs from the AttentionPooling2d layer maintain tokens and characters sequentiality, the visual embedding $I_o$ could be split into sequential region embedding tokens:
\begin{equation}
I_o=\{i_{o1},i_{o2},...,i_{oa},...,i_{oN_{2}}\},
\end{equation}
where each $i_{oa} \in \mathbb{R}^{\frac{H}{d\cdot k}\times{\frac{W}{d\cdot k}}\times{C}}$ stands for $a_{th}$ region embedding token.
Similarly, the text prompt embedding $T_p$ is divided into sequential text prompt embedding characters:
\begin{equation}
T_p=\{t_{p1},t_{p2},...,t_{pa},...,t_{pN_{2}}\},
\end{equation}
where $t_{pa} \in \mathbb{R}^{{1}\times{1}\times{C}}$ stands for $a_{th}$ the text prompt embedding character.
\par
Each region text prompt embedding $t_{pa}$ is then updated via interaction with the corresponding region embedding token $i_{oa}$:
\begin{equation}
t_{pa}=t_{pa}+l_{3}\cdot TransformerDecoder3(q=t_{pa},(k,v)=i_{oa}),
\end{equation}
where $l_3$ controls the impact of this interaction. This updates $T_p$ at the character level after encoding. 
\par
Similarly, each region embedding token $i_{oa}$ is updated via interaction with the corresponding text prompt embedding character $t_{pa}$:
\begin{equation}
i_{oa}=i_{oa}+l_{4}\cdot TransformerDecoder4(q=i_{oa},(k,v)=t_{pa}),
\end{equation}
where $l_{4}$ is also the learning parameter to control the impact of the result of character-token interaction in the equation above.
\subsection{Dual Matching Methods}
Different from the original matching mechanism in CLIP \cite{CLIP}, our proposed work introduces two parallel matching methods.
\subsubsection{Image-text Matching}
The first method image-text matching matches the text embedding $T_o$ with the visual embedding of the whole image $I_o$ to generate a global score map $S_{glo} \in \mathbb{R}^{\frac{H}{d}\times{\frac{W}{d}}\times{1}}$, formulated as below:
\begin{equation}
S_{glo}=Sigmoid(\frac{T_o I_o}{\tau}),
\end{equation}
where $\tau$ is the temperature coefficient set to 0.07.
\subsubsection{Character-token Matching}
The second method matches the text prompt embedding $T_p$ with the image embedding $I_o$ at the character-token level to produce the region score map $S_{reg}$. Each score map token $s_{a}\in \mathbb{R}^{\frac{H}{d\cdot k}\times{\frac{W}{d\cdot k}}\times{1}}$ represents $a_{th}$ part of the region score map, and is generated through the process of character-token matching as follow:
\begin{equation}
s_a=Sigmoid(\frac{t_{pa}i_{oa}}{\tau}),
\end{equation}
where $t_{pa}$ and $i_{oa}$ are the region text prompt embedding character and region embedding token with the same serial number a. Therefore, by concatenating all the $N_2$ score map tokens, the region score map $S_{reg}$ could be derived as below:
\begin{equation}
S_{reg}=\{s_1,s_2,...s_a,...,s_{N_2}\}.
\end{equation}
\subsection{Feature Enhancement and Fusion}
To achieve the final score map $S$, a mechanism is employed to enhance and fuse the global score map $S_{glo}$ and the region score map $S_{reg}$. This process involves two main steps: feature enhancement and feature fusion.
\subsubsection{Feature Enhancement}
\par
First, the region score map $S_{reg}$ could be seen as an enhancement to the global score map $S_{glo}$, capturing fine-grained features that may be overlooked by $S_{glo}$ at token level. The feature enhancement score map $S_{FE}$ is calculated as follow:
\begin{equation}
S_{FE}=S_{glo}+S_{reg}.    
\end{equation}
\subsubsection{Feature Fusion}
Second, to discover the deeper relationship between the two score maps, feature fusion is imported necessarily. This is achieved using a Transformer mechanism, which strengthens the connection between $S_{glo}$ and $S_{reg}$. The feature fusion score map $S_{FF}$ is obtained through a transformer decoder as follow:
\begin{equation}
S_{FF}=TransformerDecoder5(q=S_{glo},(k,v)=S_{reg}),
\end{equation}
where $S_{glo}$ is set to the query, while $S_{reg}$ is set to key and value.
\subsubsection{Final Score Map}
Finally, the score map $S$ is generated by combining the feature enhancement score map $S_{FE}$ and the feature fusion score map $S_{FF}$:
\begin{equation}
S=S_{FE}+\lambda_{mix}S_{FF},    
\end{equation}
where $\lambda_{mix}$ is a hyper-parameter set to 2 in this work. Then the score map $S$ is reshaped from $ \mathbb{R}^{\frac{H}{d}\times{\frac{W}{d}}\times{1}}$ to $ \mathbb{R}^{{H}\times{W}\times{1}}$ using bilinear interpolation, representing the probability of text at every pixel.
\subsubsection{Matching loss}
Similar to the contrastive loss in the original CLIP, a matching loss $L_{mat}$ is calculated for every single pixel using cross-entropy loss \cite{CEL}:
\begin{equation}
L_{mat}=\sum_{i}^{H}\sum_{j}^{W}CrossEntropy(s_{i,j},y_{i,j}),
\end{equation}
where $s_{i,j}$ is the the probability of text at pixel (i,j), and $y_{i,j}$ is the ground truth label for the presence of text at pixel (i,j). 
\par
The pixel-level score map $S$ is then fed into the downstream detection head, DBNet \cite{dbn}, which derives the final inference results for the whole task.

\subsection{Optimization}
The loss function $L_{sum}$ contains three different parts of loss function as below:
\begin{equation}
L_{sum}=L_{DB}+\lambda_{1}L_{BD}+\lambda_{2}L_{mat},
\end{equation}
where $\lambda_{1}$ and $\lambda_{2}$ are hyper-parameters set to 1 in our work, and $L_{DB}$ is the loss function of the detection head DBNet. Through the stage of training, the parameters of the whole network are updated by the reduction of the loss function $L_{sum}$.

\section{Experiment}
\subsection{Dataset}
Experiments were conducted on three renowned datasets: ICDAR2015 (IC15) \cite{ic15}, TotalText (TT) \cite{tt}, and CTW1500 (CTW) \cite{ctw}. The ICDAR2015 dataset focuses on incidental scene text and contains 1,000 training images and 500 testing images, annotated with bounding boxes and text transcriptions. The TotalText dataset emphasizes curved and arbitrarily-shaped text in natural scenes, featuring 1,255 training images and 300 testing images, annotated with polygonal shapes. The CTW1500 dataset specializes in curved text in natural scenes, comprising 1,000 training images and 500 testing images, annotated with polygonal bounding boxes to accurately depict text shapes.
\subsection{Evaluation Metric}
The performance of text detectors is evaluated using several key metrics: precision, which measures the accuracy of detected text regions; recall, which assesses the completeness of detections; F-measure, which is the harmonic mean of precision and recall; and Intersection over Union (IoU) \cite{iou}, which evaluates the overlap between detected and ground truth bounding boxes.

\subsection{Implementation Details}
In this experiment, DBNet serves as both the text detection head and a baseline for RPT, which leverages the pre-trained ResNet-50 \cite{Res} model from CLIP as its backbone. All Transformer \cite{att} decoders have 4 layers and 3 heads, with a width of 256. RPT is trained individually on the IC15, TT, and CTW datasets for 1800 epochs without pre-training and subsequently evaluated on their respective test datasets. We leverage the computing power of two TITAN Xp Graphics Cards (12GiB each) with a batch size of 4 images during training.

\begin{table}[h]
\centering
\begin{tabular}{|c|c|c|c|c|c|c|c|c|}
\hline
Method & GTP & RTP & FE & SPE & CTI & BDL & FF & IC15 \\
\cline{9-9}
       &    &    &    &    &    &   &   & F   \\
\hline
BSL   & $\times$ & $\times$ & $\times$ & $\times$ & $\times$ & $\times$ & $\times$ & 87.7 \\
\hline
BSL+   & $\checkmark$ & $\times$ & $\times$ & $\times$ & $\times$& $\times$ & $\times$ & 88.0 \\
\hline
BSL+   & $\checkmark$ & $3 \cdot 3 $ & $\checkmark$ & $\times$ & $\times$ & $\times$ & $\times$ & 88.6 \\
\hline
BSL+   & $\checkmark$ &$3 \cdot 3 $ & $\checkmark$ & $\checkmark$ & $\times$ & $\times$ & $\times$ &89.3  \\
\hline
BSL+   & $\checkmark$ &$3 \cdot 3 $ & $\checkmark$ & $\checkmark$ & $\checkmark$ & $\times$ & $\times$ & 89.8 \\
\hline
BSL+   & $\checkmark$ &$3 \cdot 3 $& $\checkmark$ & $\checkmark$ & $\checkmark$ & $\checkmark$ & $\times$ & 90.4 \\
\hline
RPT   & $\checkmark$ &$3 \cdot 3 $& $\checkmark$ & $\checkmark$ & $\checkmark$ & $\checkmark$ & $\checkmark$ & 90.5 \\
RPT   & $\checkmark$ &$4 \cdot 4 $& $\checkmark$ & $\checkmark$ & $\checkmark$ & $\checkmark$ & $\checkmark$ & 90.7 \\
\textbf{RPT}   & $\checkmark$ &$5 \cdot 5 $& $\checkmark$ & $\checkmark$ & $\checkmark$ & $\checkmark$ & $\checkmark$ & \textbf{90.9} \\
\hline
     $\Delta$&             &     &             &             & &  &  & \textbf{+3.2} \\
\hline
\end{tabular}
\caption{Ablation study of our proposed components on IC15. ``BSL'', ``GTP'', ``RTP'', ``FE'', ``SPE'', ``CTI'', ``BDL'' and ``FF''  represent the baseline method DBNet, the general text prompt, the region text prompt, the feature enhancement, and the sharing position embedding, the character-token interaction, the bidirectional distance loss, and feature fusion respectively. F (\%) represents F-measure. $\Delta$ (\%) represents the variance.}
\end{table}
\subsection{Ablation Study}
The baseline for our proposed method is established using the DBNet \cite{dbn} detection head, where the original backbone of DBNet is replaced with the ResNet-50 \cite{Res} network from the image encoder in the CLIP framework. Building on this baseline, we incrementally add the general text prompt with the fixed word embedding, region text prompt with the feature enhancement, the mechanism of sharing position embedding, character-token level interaction, bidirectional distance loss, and feature fusion to evaluate their impact on text detection performance on the IC15 dataset. The results are illustrated in Table 1.
\subsubsection{General Text Prompt with The Fixed Word Embedding}
After incorporating the general text prompt (length is 4) and fixed word embedding ``text'' into the framework, the CLIP image-text matching mechanism is activated. Compared to the baseline, this transformation significantly enhances performance by 0.3\%, illustrated in third row of Table 1.

\subsubsection{Region Text Prompt with Feature Enhancement}
With the assistance of the region text prompt, with a shape of $3 \cdot 3$, expected to be encoded in the prompt encoder, the performance of the text detection in IC15 highly boosts 0.6\%, illustrated in fourth row of Table 1, under the circumstance that character-token matching and feature enhancement mechanism is activated.
\subsubsection{Sharing Position Embedding}
Instead of directly adopting the learnable position embedding of the original CLIP text encoder, the prompt encoder utilizes the sharing position embedding to encode the region text prompt, improving the performance by 0.7\%, illustrated in fifth row of Table 1.
\subsubsection{Character-Token Level Interaction}
With the help of the interaction in character-token level before and after the encoding of the region text prompt, the performance is promoted by 0.5\%, shown in sixth row of Table 1.
\subsubsection{Bidirectional Distance Loss}
After adding the bidirectional distance loss, the performance of the text detection raises 0.6\%, shown in seventh row of Table 1. 
\subsubsection{Feature Fusion}
In contrast to simply add two score maps, the process of the feature fusion upgrades the performance of the text detection by 0.1\%, illustrated in eighth row of Table 1.
\subsubsection{The Size of Region Text Prompt}
As the size of the region text prompt increases gradually from $3\cdot 3$ to $4\cdot 4$, $5 \cdot 5$, indicating the number of characters or tokens updates from 9 to 16, 25, the performance of scene text detection improves from 90.5\% up to 90.7\%, 90.9\%. This shows that the finer-grained region text prompt could help detect text well.
\begin{table}[h]
\centering
\begin{tabular}{|c|c|c|c|c|c|c|c|c|c|c|}
\hline
\multirow{2}{*}{Method} & \multirow{2}{*}{Venue} &  \multicolumn{3}{c|}{IC15}  & \multicolumn{3}{c|}{TT}& \multicolumn{3}{c|}{CTW} \\
\cline{3-11}

& & R & P & F & R & P & F & R & P & F\\
\hline
TextSnake \cite{snake} & ECCV18 & 80.4 & 84.9 & 82.6 & 74.5 & 82.7 & 78.4 & 67.9 & 85.3 & 75.6 \\
LOMO \cite{lomo}  & CVPR19 & 83.5 & 91.3 & 87.2& 79.3 & 87.6 & 83.3 & 76.5 & 85.7 & 80.8 \\
MSR \cite{msr}  & IJCAI19 & 78.4 & 86.6 & 82.3 & 73.0 & 85.2 & 78.6  & 77.8 & 83.8 & 80.7 \\

PAN \cite{pan}  & ICCV19 & 81.9 & 84.0 & 82.9  & 81.0 & 89.3 & 85.0 & 81.2 & 84.4 & 82.7\\
DB \cite{dbn} & AAAI20 & 83.2 & 91.8 & 87.3 & 82.5 & 87.1 & 84.7 & 80.2 & 86.9 & 83.4  \\
ContourNet \cite{net}  & CVPR20 & 86.1 & 87.6 & 86.9 &  83.9 & 86.9 & 85.4 & 84.1 & 83.7 & 83.9  \\

FCENet \cite{fce}  & CVPR21 & 82.6 & 90.1 & 86.2 & 82.5& 89.3 & 85.8 & 83.4& \textbf{87.6} & 85.5\\
BPNET \cite{bpnet}  & ICCV21 & - & - & - & 84.65 & 90.27 & 87.37 & 81.45 & 87.81 & 84.51  \\
PCR \cite{pcr}  & CVPR21 & - & - & - & 82.0 & 88.5 & 85.2  & 82.3 & 87.2 & 84.7 \\
FSG \cite{tang} & CVPR22 & \textbf{90.9} & 87.3 & 89.1 &85.7 & \textbf{90.7}& \textbf{88.1}& \textbf{88.1} & 82.4 & 85.2 \\
I3CL \cite{i3cl}  & IJCV22 & - & - & - & 83.7 & 89.2 & 86.3 & 84.5 & 87.4 & \textbf{85.9} \\
DB+TCM \cite{TCM}  & CVPR23 & - & - & 89.4 & - & - & 85.9 & - & - & 85.1  \\
\hline
\textbf{RPT (Ours)} &   & 88.5 & \textbf{93.3} & \textbf{90.9} & \textbf{86.1} & 88.9 & 87.5 & 86.1 & 85.3 & 85.7 \\
\hline
\end{tabular}
\caption{Comparison with the existing paradigms of method on the dataset IC15, TT, CTW. R (\%), P (\%), F (\%) represents Recall, Precision, F-measure respectively.}
\end{table}
\subsection{Comparison with The Existing Paradigms of Method}
We compare our proposed method with the existing paradigms of the scene text detection, including TextSnake \cite{snake}, LOMO \cite{lomo}, I3CL \cite{i3cl},  MSR \cite{msr}, PAN \cite{pan}, ContourNet \cite{net}, FCENet \cite{fce}, BPNET \cite{bpnet}, PCR \cite{pcr}, FSG \cite{tang}, DB+TCM \cite{TCM} illustrated in Table 2.
From the table 2, we can conclude that
compared to existing text detection paradigms, our proposed method RPT significantly improves performance on IC15 with a SOTA performance of 90.9\% and achieves competitive performance 87.5\%, 85.7\% on TT, CTW respectively.  
\begin{figure}[h!]
    \centering
    \begin{subfigure}[b]{0.31\textwidth}
        \centering
        \includegraphics[width=\textwidth]{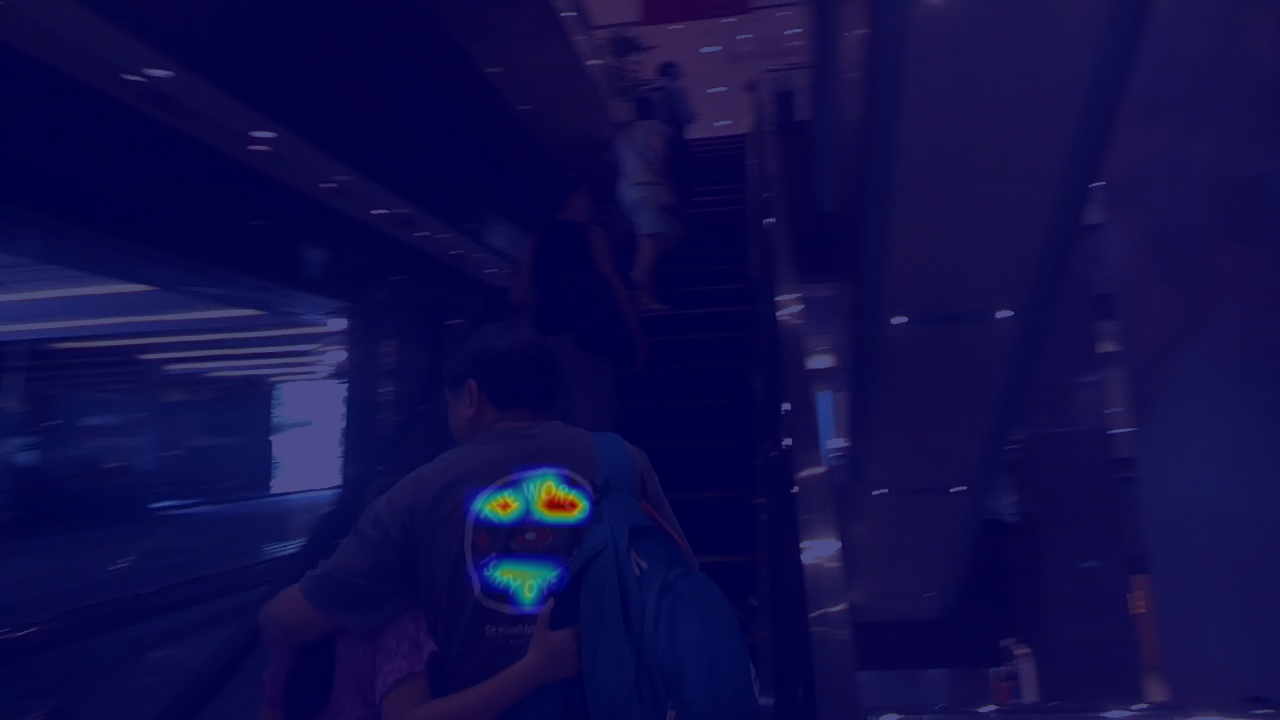}
        \caption{The heat map of sample 1's score map.}
    \end{subfigure}
    \begin{subfigure}[b]{0.31\textwidth}
        \centering
        \includegraphics[width=\textwidth]{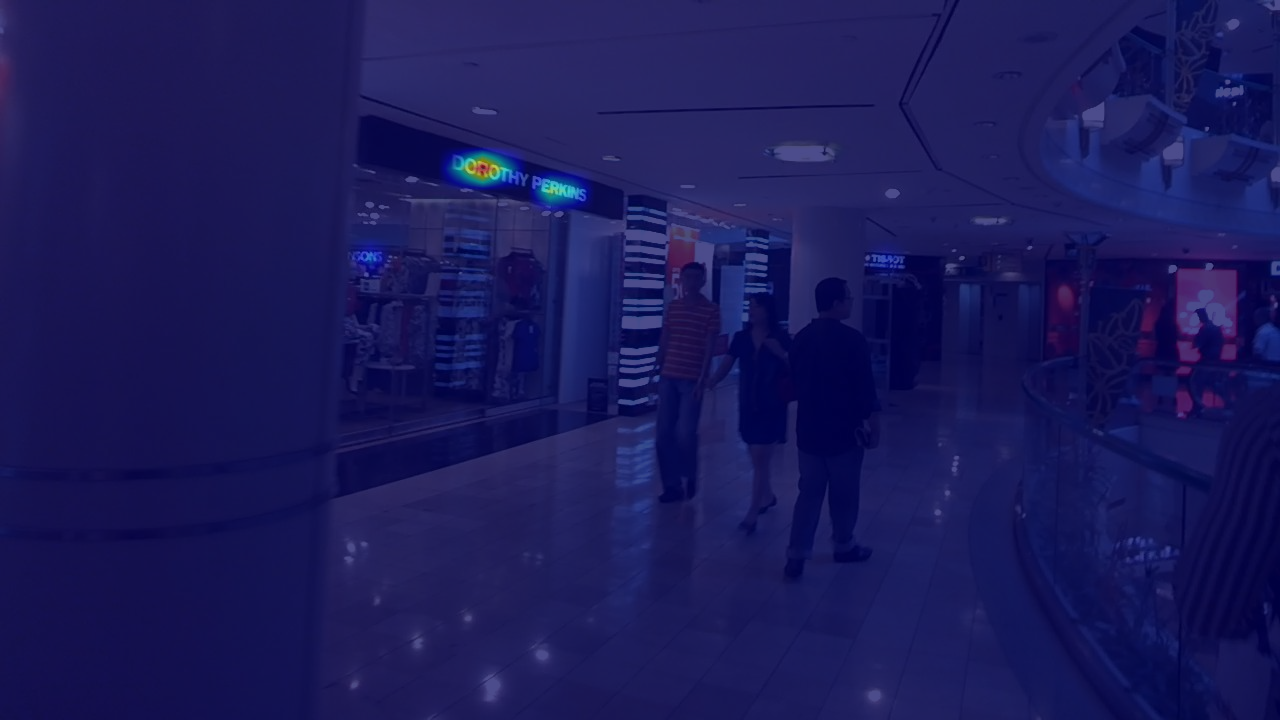}
        \caption{The heat map of sample 2's score map.}
    \end{subfigure}
    \begin{subfigure}[b]{0.31\textwidth}
        \centering
        \includegraphics[width=\textwidth]{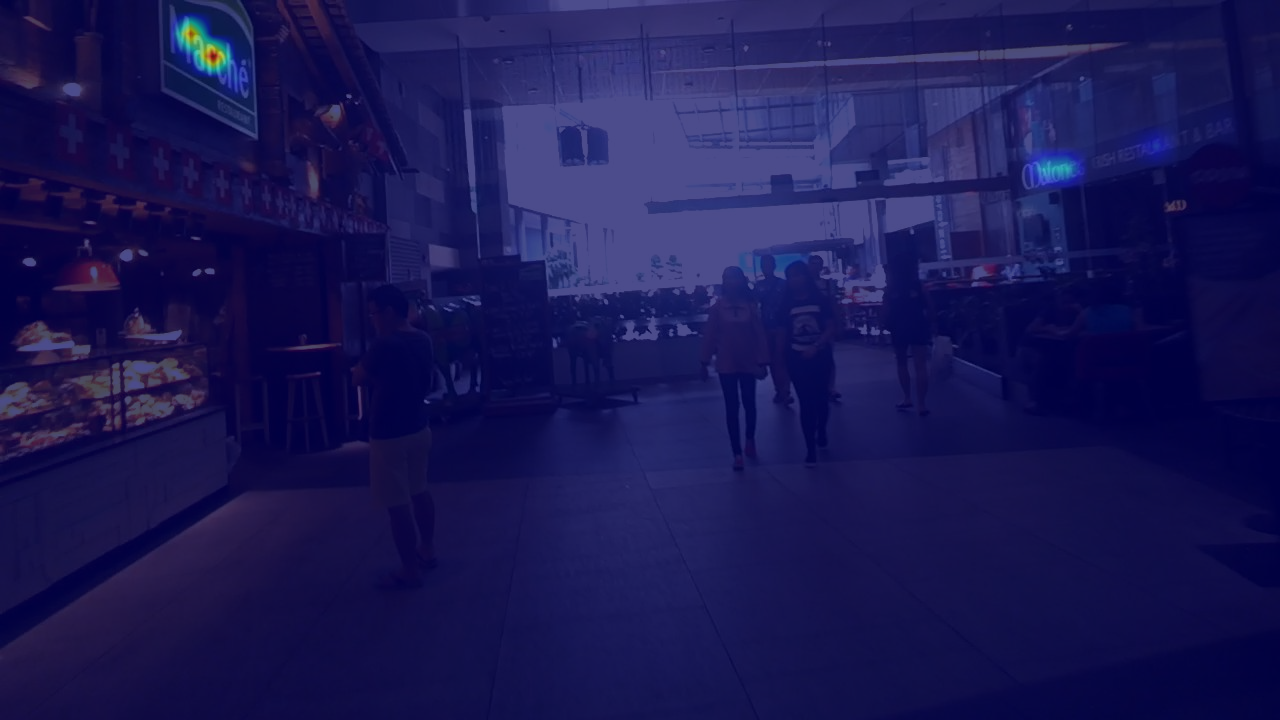}
        \caption{The heat map of sample 3's score map.}
    \end{subfigure}
    \begin{subfigure}[b]{0.31\textwidth}
        \centering
        \includegraphics[width=\textwidth]{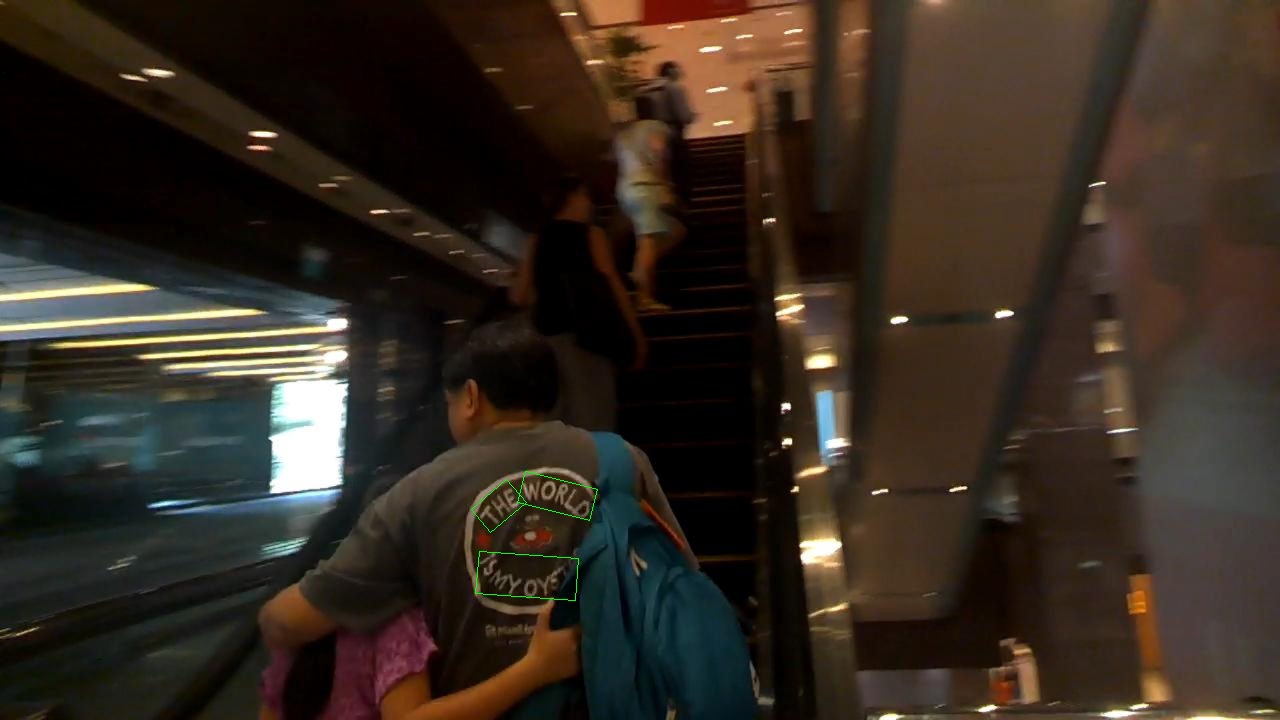}
        \caption{The detection result of sample 1.}
    \end{subfigure}
    \begin{subfigure}[b]{0.31\textwidth}
        \centering
        \includegraphics[width=\textwidth]{img_163.jpg}
        \caption{The detection result of sample 2.}
    \end{subfigure}
    \begin{subfigure}[b]{0.31\textwidth}
        \centering
        \includegraphics[width=\textwidth]{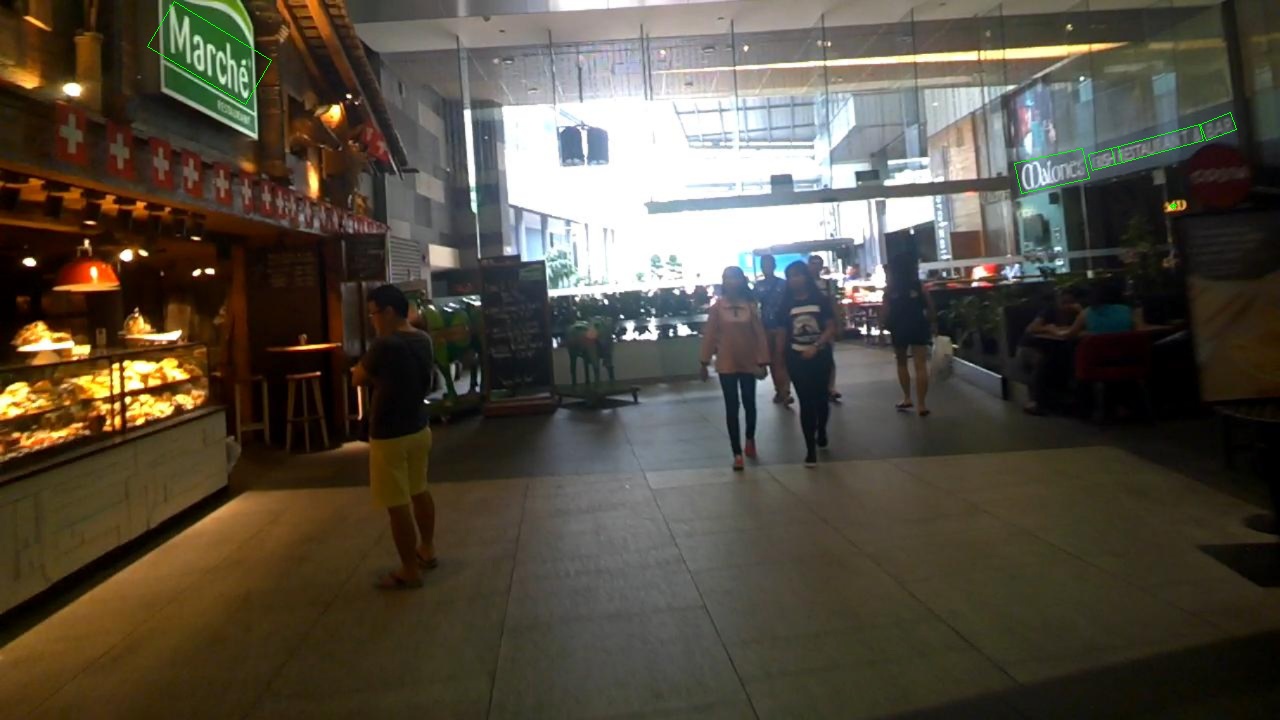}
        \caption{The detection result of sample 3.}
    \end{subfigure}
    \caption{The visualization results of 3 selected samples.}
    \label{fig:nine_images}
\end{figure}
\subsection{Visualization Results}
We selected 3 samples from the IC15 dataset, each with increasing text density. Fig.5 shows the heat map of the final score map and detection result of 3 samples respectively. In sample 1, sparse yet prominent text is accurately found by the score map shown in Fig. 5 (a), resulting in accurate detection shown in Fig. 5 (d). Sample 2, which contains moderately dense and scattered text, is identified accurately in the score map shown in Fig. 5 (b), making the detection result accurate shown in Fig. 5 (e). Sample 3 presents a very dense text, and multiple instances in close proximity. While most are detected, the current granularity of the region text prompt leads to some omissions and mistakes both in the score map shown in Fig. 5 (c) and detection result shown in Fig. 5 (f). Refining the granularity of the region text prompt could probably solve this problem because more tokens and characters can more accurately match denser text.

\section{Conclusion}
In this paper, we propose a region text tuning method (RPT) for fine-grained scene text detection, which includes the mechanism of sharing position embedding, bidirectional distance loss, character-token level interaction and feature enhancement and fusion. Extensive experiments illustrate that our proposed RPT could detect the boundaries of text in multiple challenging datasets. We would maintain further study to try different text detection heads and finer granularity of region text prompt to make region prompt tuning robuster.

\clearpage  

%
%
\bibliographystyle{splncs04}
\bibliography{main}
\end{document}